\documentclass[runningheads]{llncs}
\usepackage{graphicx}
\usepackage{hyperref}

\usepackage{amsmath}
\usepackage{amssymb}
\usepackage{caption}
\usepackage{wrapfig}
\usepackage{multirow}
\usepackage{wrapfig}

\usepackage{footnote}
\makesavenoteenv{tabular}
\makesavenoteenv{table}
\usepackage{pifont}
\newcommand{\printfnsymbol}[1]{%
  \textsuperscript{\@*}%
  }

\newcommand{\ubold}[1]{\fontseries{b}\selectfont#1}
\usepackage[export]{adjustbox}
\usepackage{array}
\usepackage{arydshln}
\usepackage{booktabs}
\newcommand{\ra}[1]{\renewcommand{\arraystretch}{#1}}
\newcommand{\comment}[1]{}
\begin{document}
\title{Importance Driven Continual Learning for Segmentation Across Domains}
%
%

\author{Sinan \"Ozg\"ur \"Ozg\"un\inst{1,2}\thanks{The authors contributed equally.} \and Anne-Marie Rickmann \inst{1\star} \and Abhijit Guha Roy\inst{1\star}
\and Christian Wachinger \inst{1} }


\authorrunning{\"Ozg\"un et al.}
%
\institute{
Artificial Intelligence in Medical Imaging (AI-Med), KJP, LMU M\"unchen, Germany \and
Computer Aided Medical Procedures, Technische Universit\"at M\"unchen, Germany
}
\maketitle              
\begin{abstract}
 The ability of neural networks to continuously learn and adapt to new tasks while retaining prior knowledge is crucial for many applications. However, current neural networks tend to forget previously learned tasks when trained on new ones, i.e., they suffer from Catastrophic Forgetting (CF). The objective of Continual Learning (CL) is to alleviate this problem, which is particularly relevant for medical applications, where it may not be feasible to store and access previously used sensitive patient data.
 In this work, we propose a Continual Learning approach for brain segmentation, where a single network is consecutively trained on samples from different domains.
 We build upon an importance driven approach and adapt it for medical image segmentation. 
 Particularly, we introduce learning rate regularization to prevent the loss of the network's knowledge. 
 Our results demonstrate that directly restricting the adaptation of important network parameters clearly reduces Catastrophic Forgetting for segmentation across domains. 

\end{abstract}
\section{Introduction}
\comment{
core dependency of neural networks on the training data DONE

not only a fixed training stage, but let the model continue to train as it is passed to new hospitals

The idea is that the model becomes more general, the higher the variety of data it has seen

However, such a sequential training of the model can lead to Catastrophic Forgetting, i.e., the networks forgets abilities it has learned before 

One possibility to circumvent CF is to mix all data in training, however, this would require that all training data is always accessible. 
Particularly in the medical field, privacy restrictions prohibit the sharing of clinical data so that a trained model may be easily passed among institutions but not the data itself. 

Methods to exactly address this issue and overcome the dilemma have been introduced in Continual Learning (CL). 
Their focus is mainly on image classification. 
In our work, we have found that a na\"ive translation to segmentation is yielding suboptimal performance. 

In this work, we study Continual Learning for brain segmentation. 
We present a novel characterization of existing methods and introduce}

After the breakthrough of Convolutional Neural Networks (CNN)~\cite{krizhevsky2012imagenet}, deep learning methods have become the technique of choice for many medical image analysis tasks such as disease classification, lesion detection, or image segmentation~\cite{litjens2017survey,ronneberger2015u}. 
Common deep architectures require large sets of training data, which are often gathered over time and across institutions. Ideally, a system would be continuously re-trained on a consolidated set of incoming data samples. However, this sequential learning requires the storage of large datasets over a long period, which can be infeasible. Particularly in the medical field, privacy restrictions can prohibit the sharing of clinical data so that a trained model may be easily passed among institutions, but not the data itself. 

Continuously fine-tuning a neural network without access to old data often results in a deterioration of the performance on prior datasets~\cite{parisi2019continual}. Neural Networks are especially prone to this phenomenon known as \emph{Catastrophic Forgetting} (CF), which emerges from the stability-plasticity trade-off~\cite{kemker2018measuring}. \emph{Continual Learning} (CL) aims to overcome this dilemma.
The focus of these methods has mainly been on classification~\cite{kirkpatrick2017overcoming,zenke2017continual}, object detection~\cite{shmelkov2017incremental}, or reinforcement learning~\cite{rusu2016progressive}. 
We have found that a na\"ive translation of such approaches to medical image segmentation is yielding sub-optimal performance. 

In this work, we explore the effect of importance driven regularization methods in an incremental domain learning setting~\cite{hsu2018re}, where at each point in time, a new magnetic resonance imaging (MRI) scan is retrieved. 
This setting poses a challenge, as neural networks can be sensitive to shifts in the input distribution that emerge from changes in the acquisition protocol, the use of different scanners, or age differences of subjects.
To overcome these problems, we define learning rate regularization 
that utilizes importance weights defined in Memory Aware Synapses (MAS)~\cite{aljundi2018memory}. 
In contrast to the soft penalty applied in MAS, we show that directly restricting the adaptation of important network parameters 
clearly reduces Catastrophic Forgetting, while preserving the ability to learn new domains.

\subsubsection{Desiderata:}
To define the Continual Learning setting and distinguish it from other learning paradigms, several desiderata have been formulated in the literature~\cite{de2019continual,farquhar2018towards,schwarz2018progress}.
The most important ones are that CL methods should be able to (i) adapt to new datasets, while retaining knowledge about old domains, (ii) without the access to old data samples, and (iii) over a long period.  Furthermore, (iv) the model should not be aware of the task or domain a data sample belongs to, i.e., the network should not have access to so-called task labels.
We strictly follow these desiderata in our Continual Learning approach for brain segmentation.

\comment{To create homogeneous settings throughout the literature, several authors have proposed~\cite{de2019continual,farquhar2018towards,schwarz2018progress} various desiderata, that should guarantee that all core ideas of CL are fulfilled. 
In our work, we implement the most important guidelines for brain segmentation. CL methods should namely be able to (i) adapt to new datasets, while retaining knowledge about old domains. (ii) This should happen over a longer period and (iii) without the access to old data samples. (iv) Furthermore, at inference time, the model should not be aware of the task or domain a data sample belongs to, i.e. the network should not have access to so-called "task labels".}
%
\subsubsection{Related Work:}
In the literature, many different techniques for CL have been proposed that can be grouped into three main categories~\cite{de2019continual}. 

\noindent
\emph{Replay-based methods} range from na\"ive rehearsal methods that store a subset of old data~\cite{rebuffi2017icarl} to pseudo-rehearsal methods that approximate previous samples using generative models~\cite{shin2017continual}.

\noindent
\emph{Parameter isolation-based methods} assign different parameters in a network to each task. This can be achieved by either fixing the architecture~\cite{mallya2018packnet}, or dynamically extending the network~\cite{xu2018reinforced}. Fixed architectures are limited by the network's capacity, whereas dynamic architectures need more memory with every new task.

\noindent
\emph{Regularization-based methods} can be divided into data-focused and prior-focused approaches. Data-focused approaches~\cite{silver2002task} distill the knowledge of old tasks to enhance the CL capabilities of the present model, whereas prior-based approaches such as \cite{kirkpatrick2017overcoming,zenke2017continual,aljundi2018memory} define importance weights for the network's parameters. Based on these weights, a regularization loss is introduced that penalizes the shift of important parameters.

\noindent
CL approaches have also been proposed in medical imaging~\cite{karani2018lifelong,baweja2018towards}, but they do not follow the desiderata formulated above. 
In~\cite{karani2018lifelong}, only batchnorm (BN) layers are fine-tuned to handle differences between domains. 
As this approach dedicates specific BN parameters to each dataset, task labels are necessary that determine to which dataset each sample belongs.
In~\cite{baweja2018towards}, the application of Elastic Weight Consolidation (EWC)~\cite{kirkpatrick2017overcoming} was proposed. The authors evaluate the method on only two consecutive tasks, in an incremental class learning setting.





\section{Methods}
\label{Methods}
\begin{figure*}[t]
  \centering
  \includegraphics[width=\linewidth]{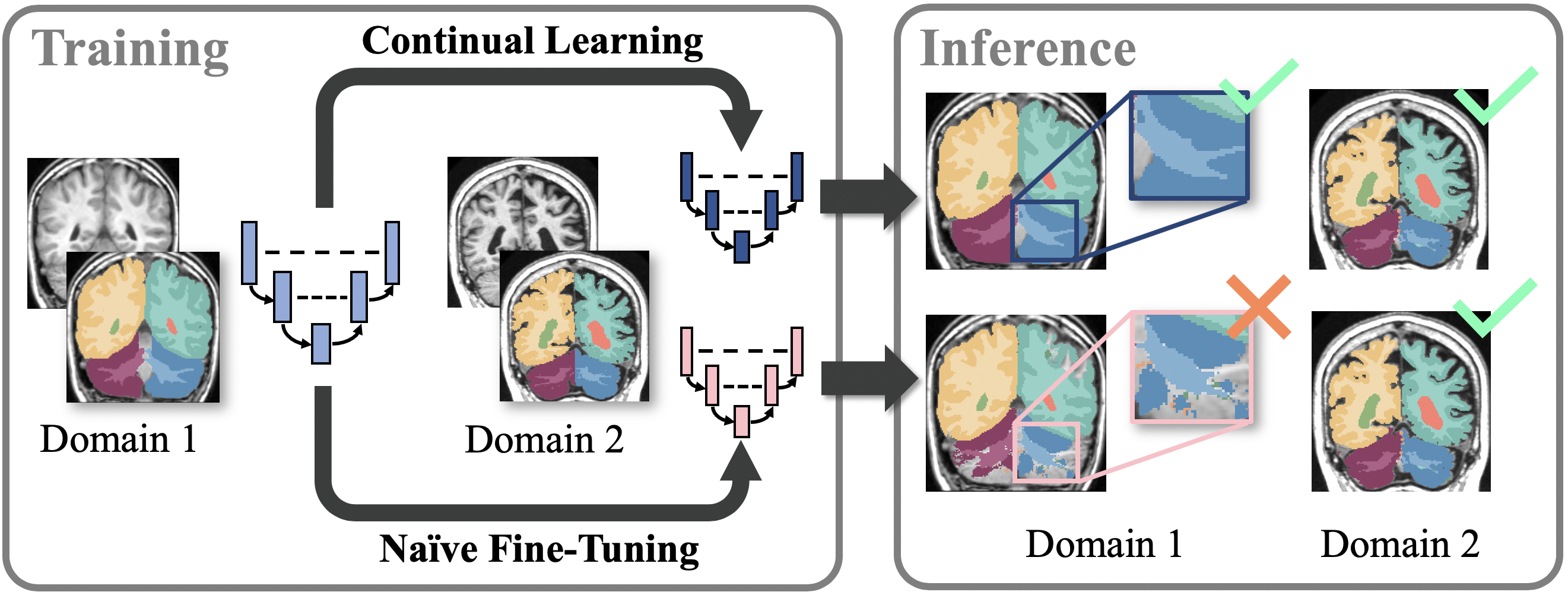}
  \caption{Illustration of Catastrophic Forgetting. Left: Segmentation network is trained on the first domain and adapted to the second using na\"ive fine-tuning (bottom) and Continual Learning (top). Right: Whereas both models adapted well to the second domain, na\"ively fine-tuning the network leads to higher CF, than CL.}
\label{fig:CF}
\end{figure*}

Our objective is to sequentially fine-tune a segmentation network on new domains without decreasing the performance on previous domains. 
Fig.~\ref{fig:CF} illustrates the effect of Catastrophic Forgetting with normal fine-tuning and the improvement with our Continual Learning approach described in the following. 

\subsection{Memory Aware Synapses}
We are building on top of Memory Aware Synapses (MAS)~\cite{aljundi2018memory}, which is one of the best performing prior-based CL methods~\cite{de2019continual}. As other prior-based CL methods~\cite{kirkpatrick2017overcoming,zenke2017continual}, MAS also defines a weight for each network parameter that reflects its importance for a specific task.
For the computation of the importance weights, MAS uses an unsupervised approach, where weights measure the sensitivity of the network's output to changes in its parameters.


\noindent
\textbf{Importance Metric:} 
Let $F$ be the learned function of the model that maps the input $X$ to the output $Y$. The importance weight $\Omega^{k}_{ij}$ for each parameter $\theta^k_{ij}$ of a 2D kernel $k$ in a convolutional layer is defined as:
\begin{equation}
    \Omega^{k}_{ij} = \frac{1}{N} \sum_{n=1}^{N} \frac{\partial||F(x_n;\theta)||^2_2}{\partial \theta^{k}_{ij}},
\end{equation}
where $\frac{\partial||F(x_n;\theta)||^2_2}{\partial \theta^k_{ij}}$ is the gradient of the squared $\ell2$ norm of the softmax output for data point $x_n$ with respect to the parameter $\theta^{k}_{ij}$ and $N$ is the number of samples. The importance weights are computed after training on each domain and we accumulate a moving average over all tasks.

\noindent
\textbf{Surrogate Loss:}
MAS employs the importance weights by introducing an additional surrogate loss that penalizes the change of important network parameters. The total loss function is then defined as:
\begin{equation}
    L(\theta) = L_d(\theta) + \lambda \sum_{kij} \Omega^{k}_{ij}(\theta^{k}_{ij} - \theta^{*k}_{ij})^2,
    \label{eq:Masreg}
\end{equation}
where $L_d$ describes the regular loss of the current domain, $\Omega$ the important weights, $\theta$ the current network parameters and $\theta^*$ the parameters of the old network, trained on the previous task. The hyperparameter $\lambda$ controls the impact of the regularization term on the loss function. 

\noindent
\textbf{Adaptation of MAS to brain segmentation:}
The original surrogate loss in Eq.~(\ref{eq:Masreg}) is dependent on the size of the network, which makes it unstable for large networks. 
Consequently, we divide the surrogate loss by the number of network parameters.
Moreover, directly applying the importance calculation of MAS to brain segmentation was not feasible, as highly skewed importance weights resulted in an unstable training process. 
To counter this problem, we detected outliers based on the interquartile range criterion and set them to their respective boundaries. 
We normalized the resulting importance values between zero and one to increase their interpretability. 

\subsection{Learning Rate Regularization} 

In this section, we propose an alternative regularization approach for MAS. 
In contrast to regularizing changes of important parameters using a surrogate loss, we define a parameter-specific learning rate such that the learning rate becomes a function of a parameter's importance~\cite{ebrahimi2019uncertainty}. 
Hence, the learning rate of important parameters will be reduced, while the learning rate for non-important parameters will be kept the same.
While the surrogate loss in MAS indirectly penalizes changes of the parameters, learning rate regularization provides a more direct means to avoid changes to important parameters. 
Let $\alpha^d$ be the learning rate for training domain $d$, the parameter-specific learning rate is:
\begin{equation}
    \alpha^{d}_{kij} = (1-\Omega^{k}_{ij}) \alpha^{d}.
    \label{equ:maslr}
\end{equation}
We refer to this approach as \textbf{MAS-LR}. Another advantage is its lack of additional hyperparameters.

\subsubsection{Parameter Freezing:}
A soft penalty might not be sufficient to enforce the network to retain old knowledge. 
Taking inspiration from network pruning methods~\cite{mallya2018packnet,li2016pruning}, we propose \textbf{MAS-Fix}, which freezes important parameters during the training of a new task and only fine-tunes unimportant parameters. 
This approach can be interpreted as a hard regularization of Eq.~(\ref{equ:maslr}) by either setting $\alpha^{d}_{kij} = \alpha^{d}$ or $\alpha^{d}_{kij} = 0$. 
We define hyperparameters $\beta_d$ for each domain~$d$ that define a threshold of how much of the network can be frozen in each step. 
If all important parameters of a succeeding task were already fixed in earlier stages, the network will remain as is. In doing so, we do not enforce the network to freeze unimportant parameters and thus extend the remaining capacity of the network and the number of forthcoming tasks.

\subsubsection{Filter and Kernel Importance:} MAS defines importance weights for each parameter of a network. This can be cumbersome for larger networks, as it doubles the amount of memory needed to save the model.
In our experiments, we observed that importance values within a convolutional kernel are similar, which is expected due to the shared functionality of the parameters in a kernel. In addition, values within a filter also had similar values, even though the similarity within a kernel was higher.
Hence, it would be meaningful to assign importance weights on the level of kernels or filters, instead of single parameters. 
The filter and kernel importance can be easily integrated into the proposed regularization of the learning rate by  averaging all weights within a kernel or filter, respectively.


\subsection{Evaluation Metrics}
\begin{figure*}[t]
  \centering
  \includegraphics[width=\linewidth]{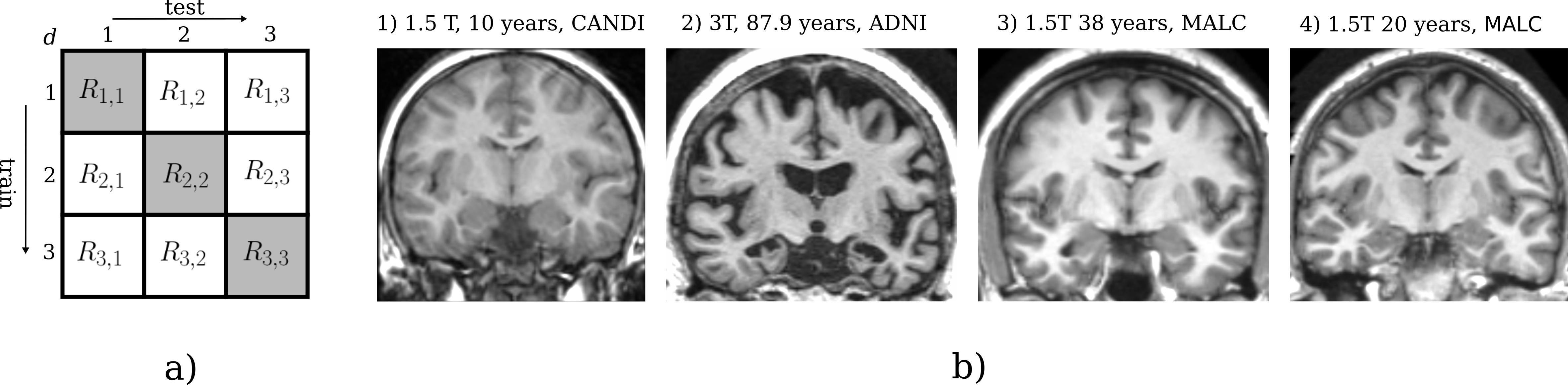}
  \caption{a) Train-Test-Accuracy-Matrix. b) Slices of brain scans from the four different domains with magnetic field strength, age and original dataset. }
\label{fig:domains}
\end{figure*}

\comment{
\begin{wrapfigure}[h]
    \centering
     \includegraphics[width= 0.30 \textwidth]{images/metrics_table_3.png}
    \caption{Train-Test-Accuracy matrix}
    \label{fig:metrics_table}
\end{wrapfigure}}
To evaluate CL methods, it is important to not only focus on the improvement of Catastrophic Forgetting but also to consider the accuracy over time and knowledge transfer to unseen domains.
Hence, we adopt some of the metrics proposed in~\cite{diaz2018don}.
Central is the train-test-accuracy-matrix, $R\in \mathbb{R}^{D \times D}$, illustrated in Fig.~\ref{fig:domains}a,
where $D$ determines the number of domains and each entry $R_{i,j}$ is the mean Dice score (DSC) of the model on domain $j$ after training on domain~$i$~\cite{lopez2017gradient}.

\noindent
\textit{Transfer Learning} (TL) is the average of the diagonal of $R$ and measures the plasticity, i.e., the ability to adapt to new tasks. 
\textit{Backward Transfer} specifies the effect that learning a new task has on the performance on older tasks. The performance can either degrade, in case of CF, or ideally increase. BWT is, therefore, broken down into its two components, REM and $\text{BWT}^+$. 
We slightly modify the metrics introduced in~\cite{lopez2017gradient}, as shown in Eq.~\ref{eq:rem} and Eq.~\ref{eq:bwt}.
\textit{Remembering} (REM) measures the stability of the model, i.e., the ability of the network to retain its knowledge. This metric assesses the effect of a CL method on Catastrophic Forgetting
\begin{equation}
    \label{eq:rem}
     REM = \frac{2\sum_{i=2}^{D}\sum_{j=1}^{i-1} 1 - \left | \min{(R_{i,j} - R_{j,j}, 0)} \right |}{D(D-1)}.
\end{equation}
\textit{Positive Backward Transfer} ($\text{BWT}^+$) measures the improvement of the network on old domains by accommodating new knowledge
\begin{equation}
    \label{eq:bwt}
    BWT^{+} = \frac{2\sum_{i=2}^{D}\sum_{j=1}^{i-1} \max{(R_{i,j} - R_{j,j}, 0)}}{D(D-1)}.
\end{equation}
\textit{CL Dice Score} (CL DSC) combines the transfer learning and backward transfer abilities of the network and is the most generic metric to evaluate Continual Learning. This metric is calculated as the average of the entries of the diagonal and below the diagonal of $R$.
\textit{Forward Transfer} (FT) is the average performance of the network on unseen domains (entries above the diagonal). This metric does not explicitly measure the Continual Learning abilities of the system, but it is an essential indicator of the network's ability to generalize well on unseen data. 

\comment{
\begin{figure}[h]
  \centering
  \includegraphics[width=\linewidth]{images/metrics_figure.png}
\label{fig:metrics}
\end{figure}}
\comment{
\begin{align}
        TL &= \frac{\sum_{i}^{D} R_{i,i}}{D} \\
        REM &= \frac{\sum_{i=2}^{D}\sum_{j=1}^{i-1} 1 - \left | \min{(R_{i,j} - R_{j,j}, 0)} \right |}{\frac{D(D-1)}{2}}\\
        BWT^{+} &= \frac{\sum_{i=2}^{D}\sum_{j=1}^{i-1} \max{(R_{i,j} - R_{j,j}, 0)}}{\frac{D(D-1)}{2}}\\
        FWT &= \frac{\sum_{i<j}^D R_{i,j}}{\frac{D(D-1)}{2}}\\
       CL_{acc} &= \frac{\sum_{i\geq j}^{D} R_{i,j}}{\frac{D(D+1)}{2}} 
\end{align}}

\section{Experiments and Results}

\comment{
\begin{figure*}[h]
  \centering
  \includegraphics[width=\linewidth]{images/segmentation_matrix.png}
  \caption{}
\label{fig:segmentations}
\end{figure*}}
\subsection{Experimental Setup}
\subsubsection{Datasets:}
We use MRI T1 brain scans from three different datasets with manual segmentations: Child and Adolescent NeuroDevelopment Initiative (CANDI)~\cite{kennedy2012candishare}, Multi-Atlas Labelling Challenge (MALC)~\cite{landman2012miccai}, and Alzheimer's Disease Neuroimaging Initiative (ADNI)~\cite{jack2008alzheimer}.
The datasets differ in their age ranges, MRI field strengths and presence of pathologies or motion artifacts. All datasets were re-sampled to have  an  isotropic  resolution  of 1mm$^3$. 
\comment{Manual segmentations of all brain scans were provided by Neuromorphometrics, Inc.}
\comment{
The Multi-Atlas Labelling Challenge(MALC) dataset~\cite{landman2012miccai}  is  part  of  the  OASIS  dataset~\cite{marcus2010open}.  It consists  of 30 MRI  scans of different subjects. It covers an age range from 18 to 90 years.
ADNI-29 Dataset: The  dataset  consists  of  29  scans from  the  ADNI  dataset~\cite{jack2008alzheimer},  with  a  balanced  distribution  of Alzheimer’s Disease and control subjects, and scans acquired with 1.5T and 3T scanners. CANDI Dataset: The  dataset  consists  of  13  brain scans  of  children  (age  5-15)  with  psychiatric  disorders  and is  part  of  the  CANDI  dataset~\cite{kennedy2012candishare}.  Some  scans  have  severe motion artifacts.}

\subsubsection{Domains:}
We adopt an online learning approach, where at each point in time, a new volume is used to update the network. 
For our setting, we define four different domains that we train on consecutively. We assume that in order to build a working system, it has to be initially trained on a larger dataset. Therefore, we train the network from scratch on our first domain, which consists of 13 scans from the CANDI dataset. The next domains contain one training scan from other datasets and differ from the first domain in age range, field strength, or presence of pathology. Fig.~\ref{fig:domains} shows the slices of brain scans of the four different domains. Noticeable is, for example, the ringing artifacts in the first domain, differences in intensities and enlarged ventricles in the second domain. 
We chose this setting, as we assume that volumes are collected one at a time because annotating medical scans is not only time-consuming but also costly.

\subsubsection{Network Architecture and Training Parameters:}
We choose QuickNAT~\cite{roy2019quicknat} as our baseline architecture as it achieves state-of-the-art performance in brain segmentation. 
For simplicity, we do not perform view aggregation as proposed in~\cite{roy2019quicknat} and only train on coronal slices. 
We train the networks using stochastic gradient descent with momentum of $0.95$ for 12 epochs on each domain. We choose an initial learning rate of $0.1$ and reduce it every 4 epochs by a factor of $0.5$ during training of the first domain. For all succeeding domains, the learning rate is not further reduced, unless specified in the parameter-specific learning rates training scheme. For our CL methods, we conduct a hyperparameter search where necessary. To balance the surrogate loss proposed in the original MAS method, we set $\lambda$ to $10^4$. In MAS-Fix, we set $\beta_d$ to 0.25 for $d \in [1, D]$ , to freeze 25\% of the network after the first stage. For every consecutive training step, we allow the model to fix up to additional 25\%. MAS-LR does not require an additional hyperparameter. 
Based on results on the validation sets, we use kernel level importance weights for MAS-Fix and filter level importance weights for MAS-LR. For MAS we did not observe an improvement for aggregating weights.

\comment{ 
\noindent
\textbf{Baselines:}
We compare to na\"ive fine-tuning of the network (lower bound) and joint training of the network on a combined dataset (upper bound for REM and $\text{BWT}^+$) as baselines.
Many modern networks are trained using regularization techniques like $\ell2$ regularization and dropout~\cite{srivastava2014dropout}, which can reduce Catastrophic Forgetting~\cite{goodfellow2013empirical,aljundi2018memory,kirkpatrick2017overcoming}, therefore we also compare to models using these techniques. We want to emphasise that all other models do not use dropout or $\ell2$ regularization, as this complicates the evaluation of CL methods.
}
\begin{table}[t]
    \centering
    \caption{Comparison of baselines (top) and our importance based training methods (bottom). The CL metrics are calculated on the average Dice scores over all segmented structures.}
     \ra{1.3} 
    \resizebox{\linewidth}{!} {
    \begin{tabular}{@{}l l c @{\hskip 5\tabcolsep} c @{\hskip 5\tabcolsep} c @{\hskip5         \tabcolsep} c @{\hskip 5\tabcolsep} c@{}}
        \toprule
         &Method & \includegraphics[width=15pt,valign=c]{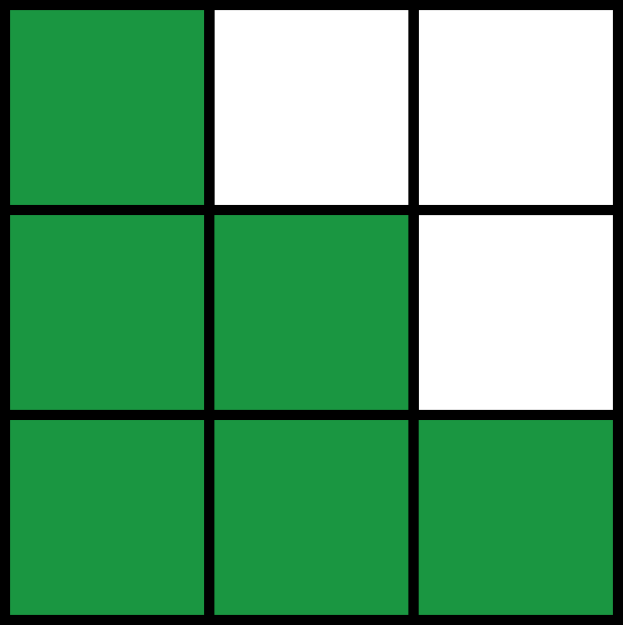}CL DSC &  \includegraphics[width=15pt,valign=c]{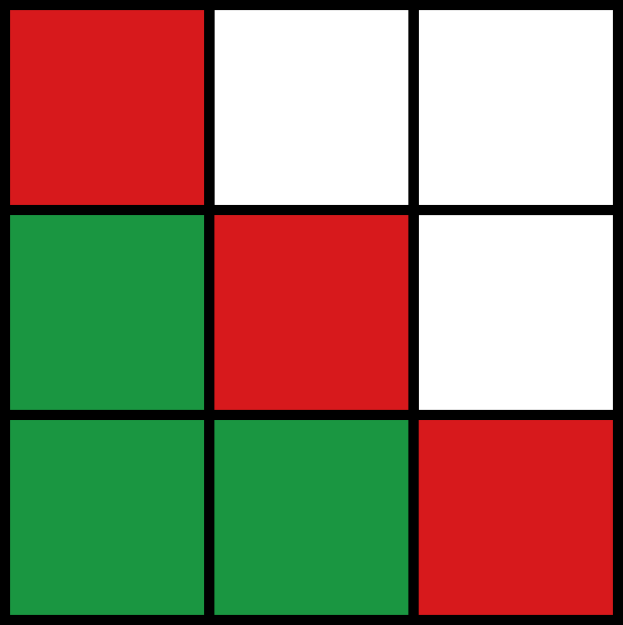}REM &  \includegraphics[width=15pt,valign=c]{images/bwt.png}$\text{BWT}^+$ & 
         \includegraphics[width=15pt,valign=c]{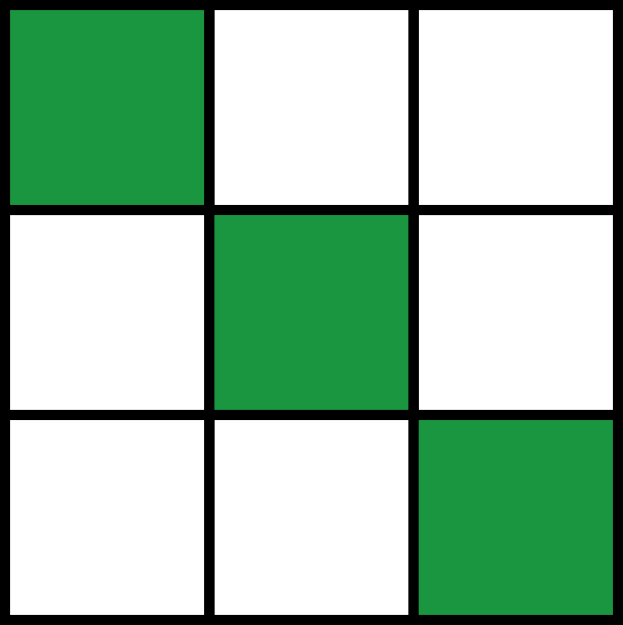}TL &
         \includegraphics[width=15pt,valign=c]{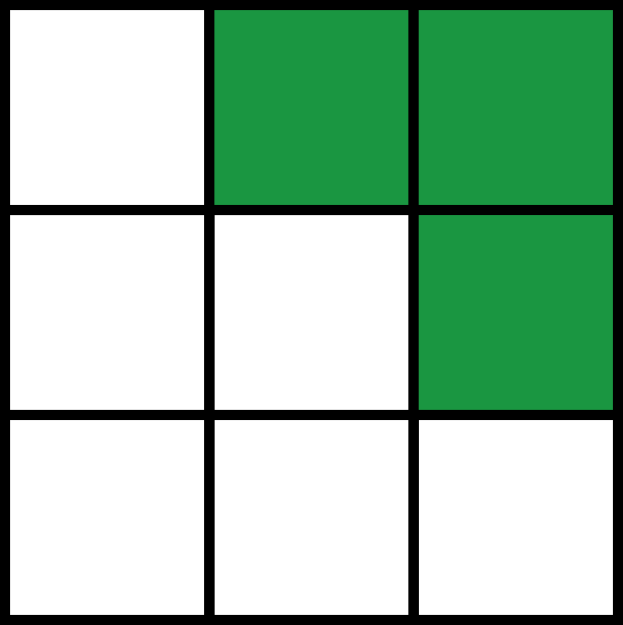}FWT \\
        \midrule
        Baseline & 
        Fine-tuning & $0.716 \pm 0.138$ & $0.903 \pm 0.054$ & $0.010 \pm 0.016$ & $0.770 \pm 0.118$& $0.605 \pm 0.174$\\

        &Joint training & $0.783 \pm 0.108$& $0.997 \pm 0.003$ & $0.029 \pm 0.024$ & $0.776 \pm 0.114$& $0.650 \pm 0.161$\\[0.5em]
        \midrule
        CL Methods & MAS
        & $0.742 \pm 0.129$& $0.929 \pm 0.045$ & $0.007 \pm 0.011$& $0.783 \pm 0.112$ & $0.625 \pm 0.166$\\
        &MAS-Fix & $0.755 \pm 0.130$& $0.954 \pm 0.036$ & $0.010 \pm 0.014$& $0.781 \pm 0.116$ & $0.634 \pm 0.166$\\
         &MAS-LR & $0.756 \pm 0.128$& $0.960 \pm 0.032$ & \ubold{0.012 $\pm$ 0.016} & $0.778 \pm 0.117$& $0.637 \pm 0.164$\\
         \midrule
        Regularization &$\ell2$ & $0.726 \pm 0.135$& $0.913 \pm 0.054$ & $0.011 \pm 0.012$ & $0.778 \pm 0.108$& $0.614 \pm 0.185$\\
        &Dropout  & $0.744 \pm 0.146$&$0.931 \pm 0.055$ & $0.007 \pm 0.008$ & \ubold{0.786 $\pm$ 0.121}& $0.690 \pm 0.169$\\
         \midrule
         Combination &MAS-LR-Dropout & \ubold{0.761 $\pm$ 0.132} & \ubold{0.965 $\pm$ 0.026} & $0.009 \pm 0.008$ & $0.778 \pm 0.124$& \ubold{0.693 $\pm$ 0.170} \\

        \bottomrule
    \end{tabular}}
    
    \label{tab:main_results}
\end{table}
\subsection{Results}
We compare our methods to na\"ive fine-tuning of the network (lower bound) and joint training of the network on a combined dataset (upper bound for REM and $\text{BWT}^+$) as baselines and report the results in Tab.~\ref{tab:main_results}. 
We observe an increase in the overall CL Dice score for all CL methods, where learning rate regularization (MAS-LR and MAS-Fix) performs better than regularization using a surrogate loss as proposed in MAS. 
As both of our methods provide the ability to set the learning rate to zero for important parameters, they lead to higher stability (better remembering), while the ability to learn new knowledge is comparable to MAS.
Interestingly, in contrast to the stability-plasticity trade-off, we observe not only a better Remembering (REM) but also an increase of the Transfer Learning (TL) performance for all CL methods.
We believe this is due to the regularization effect of the methods, that help the model to generalize better and thus achieve higher performance. 
The ability to generalize well on unseen data (FWT), also increases using regularization techniques.
Positive Backward Transfer ($\text{BWT}^+$) does not differ much between the CL methods. MAS leads to a slight decrease in $\text{BWT}^+$, which could be caused by the higher Transfer Learning capability. In general, Positive Backward Transfer is hard to achieve, as even the upper bound has a low score in this metric.
We show segmentation results in Fig.~\ref{fig:segmentations}, where we compare the effect of na\"ive fine-tuning and MAS-LR on Backward Transfer, Forward Transfer and Transfer Learning for the first two domains. 

As many modern networks are trained using regularization techniques like $\ell2$ regularization and dropout~\cite{srivastava2014dropout}, which can reduce Catastrophic Forgetting~\cite{goodfellow2013empirical,aljundi2018memory,kirkpatrick2017overcoming}, we also compare to models using these techniques.  We observe a slight improvement using $\ell2$ regularization, whereas dropout even outperforms MAS. Dropout specifically leads to the highest increase in Forward Transfer. 
Finally, to determine how dropout influences CL methods, we trained a network with dropout and our best performing CL-method (MAS-LR), which further increased the Remembering and FWT performance.

\begin{figure}[t]
  \centering
  \includegraphics[width=\linewidth]{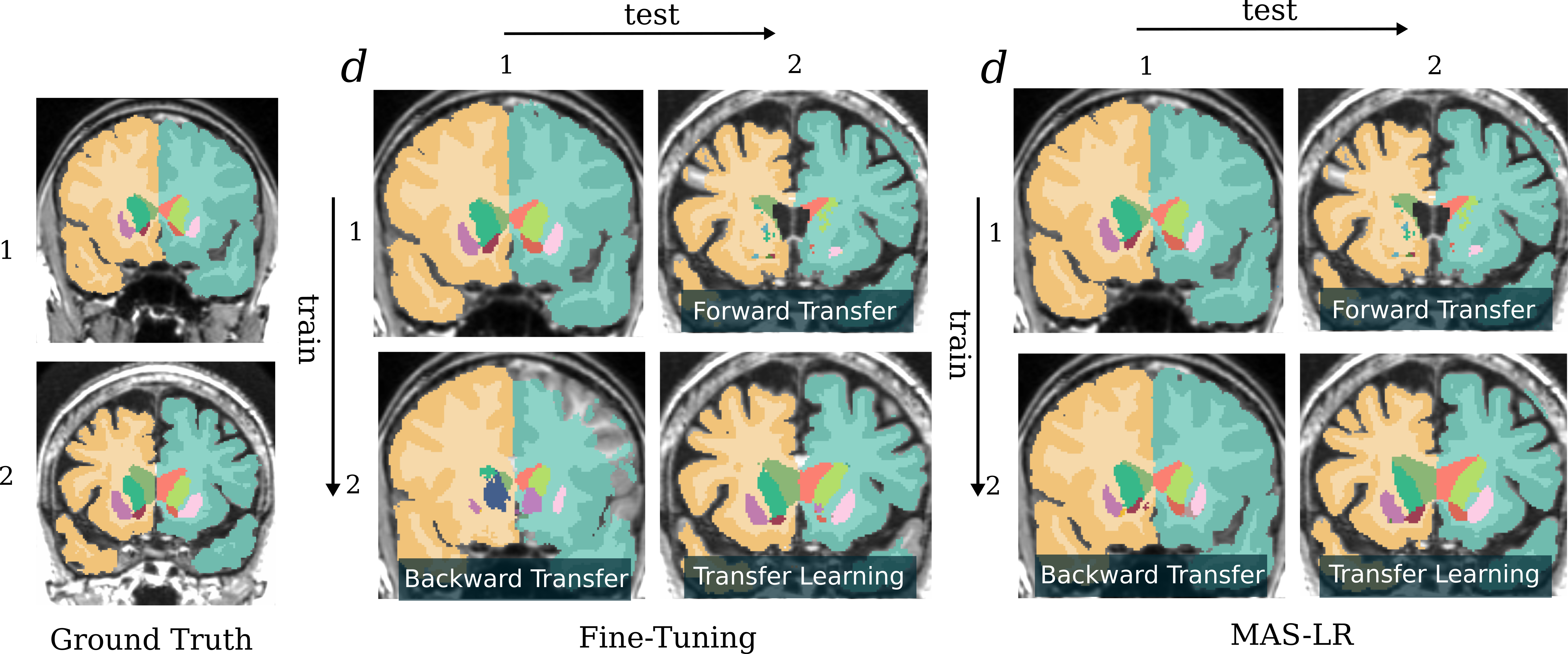}
  \caption{Comparison of segmentation results on the first two domains for the na\"ive fine-tuning baseline and our proposed method MAS-LR. Most striking are the improvements in Backward Transfer. }
\label{fig:segmentations}
\end{figure}

\section{Conclusion}
We proposed an Importance Driven Continual Learning approach for brain segmentation across domains. We adapted the importance weights introduced in MAS to our medical setting. 
We observed that detecting outliers in the importance weights and normalizing them, leads to a more stable training and higher performance.
In contrast to the surrogate loss used in MAS, we proposed learning rate regularization to restrain changes to important network parameters.
MAS-LR outperformed MAS by clearly reducing catastrophic forgetting, without the need for additional hyperparameters. We further demonstrated that learning rate regularization can be combined with standard regularization approaches like dropout.

\comment{
\section*{Acknowledgements}
This research was partially supported by the Bavarian State Ministry of
Science and the Arts in the framework of the Centre Digitisation.Bavaria (ZD.B). We thank NVIDIA Corporation for GPU donation.}

\vspace{-0.3cm}
\bibliographystyle{splncs04}
\bibliography{references}

\end{document}